\documentclass[conference,letterpaper]{IEEEtran}
\IEEEoverridecommandlockouts
\usepackage[noadjust]{cite}
\usepackage{amsmath,amssymb,amsfonts}
\usepackage{graphicx}
\usepackage{textcomp}
\usepackage{xcolor}
\usepackage{booktabs}
\usepackage{array}
\usepackage{xurl}
\usepackage{multirow}
\usepackage{algorithm}
\usepackage{algpseudocode}
\usepackage[hidelinks,hypertexnames=false]{hyperref}

\def\BibTeX{{\rm B\kern-.05em{\sc i\kern-.025em b}\kern-.08em
    T\kern-.1667em\lower.7ex\hbox{E}\kern-.125emX}}

\renewcommand{\figurename}{Figure}
\renewcommand{\tablename}{Table}
\renewcommand{\thetable}{\arabic{table}}
\DeclareRobustCommand{\supercite}[1]{\unskip\textsuperscript{\cite{#1}}}
\begin{document}

\title{PAC-ACT: Post-training Actor-Critic for Action Chunking Transformers}

\author{
\IEEEauthorblockN{Yujie Pang}
\IEEEauthorblockA{\textit{Dept. of Mechanical and}\\
\textit{Energy Engineering}\\
\textit{Southern University of}\\
\textit{Science and Technology}\\
Shenzhen 518055, China\\
0009-0003-6377-6949}
\and
\IEEEauthorblockN{Zudong Li}
\IEEEauthorblockA{\textit{Dept. of Mechanical and}\\
\textit{Energy Engineering}\\
\textit{Southern University of}\\
\textit{Science and Technology}\\
Shenzhen 518055, China\\
0009-0009-5574-9602}
\and
\IEEEauthorblockN{Leran Chen}
\IEEEauthorblockA{\textit{Dept. of Mechanical and}\\
\textit{Energy Engineering}\\
\textit{Southern University of}\\
\textit{Science and Technology}\\
Shenzhen 518055, China\\
chenlr@sustech.edu.cn}
\and[\hfill\mbox{}\par\vspace{0.8em}\mbox{}\hfill]
\IEEEauthorblockN{Yongsheng Ma\textsuperscript{*}%
\thanks{\textsuperscript{*}Corresponding author: Yongsheng Ma (\nolinkurl{mays@sustech.edu.cn}).}}
\IEEEauthorblockA{\textit{Dept. of Mechanical and}\\
\textit{Energy Engineering}\\
\textit{Southern University of}\\
\textit{Science and Technology}\\
Shenzhen 518055, China\\
mays@sustech.edu.cn}
}

\maketitle

\begin{abstract}
Precision industrial contact manipulation requires robots to complete tasks reliably under pose perturbations and contact-force constraints, placing strong demands on policy determinism, contact-force safety, and disturbance resistance. Vision-language-action models provide strong task generalization, but they usually introduce higher inference and GPU-memory overhead, which still limits deployment in high-frequency control scenarios. In contrast, vision-action chunking policies provide low-latency inference and temporally continuous actions, making them more suitable for industrial scenarios that require predictable and real-time behavior. However, these policies are mainly trained through behavior cloning, and error accumulation under distribution shift limits further improvements in task success and contact stability. To address this issue, this paper proposes PAC-ACT, a reinforcement-learning post-training framework for pretrained action-chunking policies. The core idea is to improve task completion and contact stability through reinforcement learning while preserving the lightweight and efficient properties of vision-action chunking models. Specifically, PAC-ACT reformulates step-wise policy optimization as a chunk-level decision process, aligning reinforcement-learning updates with action-chunk generation; it further builds policy and value-network architectures adapted to pretrained networks and introduces a hybrid behavior-prior constraint to prevent the policy from drifting away from the action distribution learned by the pretrained policy. Experiments on industrially motivated precision-contact benchmarks show that the method improves task success, contact stability, and force safety while maintaining low latency and low GPU-memory usage. In particular, on the Contour task, the median peak contact force is significantly reduced, and the proportion of force readings above 60N is reduced by 46 times. An additional sparse-reward ablation verifies the necessity of key framework components and shows that PAC-ACT can still explore effectively under randomized initial poses when only the task-success reward is retained together with the behavior-prior constraint.
\end{abstract}


\section{Introduction}

Model-based or vision-servoed manipulation pipelines often depend on accurate camera calibration, hand-eye calibration, and object or end-effector pose estimation. Errors in these calibration and localization steps can propagate through the visual-servoing control law and degrade positioning accuracy, which becomes particularly problematic in precision contact tasks \supercite{bVisualServo,bMeasurementError}. In recent years, end-to-end visuomotor policies represented by Action Chunking Transformer (ACT) and Diffusion Policy have directly mapped high-dimensional visual observations to low-dimensional actions, reducing the need for explicit geometric action design in several manipulation settings \supercite{b1,b2}. In particular, ACT predicts multiple future action steps at once through action chunking, improving temporal continuity and execution stability as reported in fine-grained bimanual manipulation experiments \supercite{b1}. This paradigm has been extended to bimanual manipulation \supercite{b1}, multi-task robot learning \supercite{b23}, and large-scale vision-language-action (VLA) models \supercite{b18,b10}, demonstrating the broad applicability of action chunking in robot control.

However, such action chunking policies are still fundamentally trained under the offline behavior cloning paradigm, and their performance strongly depends on the quality and coverage of demonstrations \supercite{b1,b2,b9}. When the robot encounters out-of-distribution states or contact perturbations at test time, small action deviations may accumulate over long-horizon execution and eventually cause failure, a common limitation of behavior cloning under covariate shift \supercite{bDAgger,b9}. This distribution-shift problem is particularly severe in precision contact tasks: millimeter-level pose errors may either prevent the robot from reaching the target contact region or drive the contact force beyond the safety range. Studies on contact manipulation further note that sustained in-contact tasks generally require explicit or implicit regulation of robot-environment interaction forces \supercite{bContactSurvey,bImpedance}. Therefore, action chunking policies trained only by offline behavior cloning remain insufficient for industrial contact tasks that require out-of-distribution robustness and force safety \supercite{bDAgger,b9,bContactSurvey,bImpedance}.

Directly introducing action chunking policies into reinforcement learning faces structural challenges. The mismatch between chunk-level action generation and step-wise reward feedback exacerbates credit assignment, motivating recent studies on chunk-level RL and action-sequence value estimation \supercite{b5,b6,b14}. Existing studies have begun to explore chunk-level RL, including Chunking the Critic \supercite{b5} and Q-Chunking \supercite{b6}, but they mainly focus on learning chunking policies from scratch or improving offline-to-online transfer efficiency. The problem of reusing the structure of a pretrained action chunking policy while preserving its behavior prior during fine-tuning remains underexplored.

To fill this gap, we propose PAC-ACT, a reinforcement-learning post-training framework for pretrained ACT policies. Built on a pretrained ACT model under the LeRobot \supercite{b8} framework, PAC-ACT exploits the action prior learned from a moderate number of expert demonstrations and reduces the dependence of online RL on large-scale random exploration. Specifically, we design an ACT-transferred Actor-Critic architecture that preserves the ACT backbone while removing the CVAE latent module, introduce hybrid KL constraints to maintain the pretrained behavior prior, and reformulate the MDP to resolve the structural mismatch between step-wise RL and chunk-level action generation.

The main contributions are summarized as follows:
(1) We propose an RL post-training framework for pretrained ACT policies and construct an Actor-Critic architecture from the native ACT structure to enable stable online fine-tuning.
(2) We jointly improve force safety and task completion efficiency for industrial precision contact tasks, enhancing contact stability and execution efficiency while retaining low-latency inference.
(3) We conduct systematic experiments on Metal Touch and Square Assembly, providing detailed quantitative analyses of task success, force-safety improvement, and training convergence dynamics, with significant reductions in hazardous force events and extreme peak forces on the Contour task.
(4) We demonstrate the potential of RL post-training for task-specific optimization in industrial robotics and show that the behavior prior learned by behavior cloning can effectively reduce the exploration burden of reinforcement learning.

\section{Related Work}

\subsection{Transformer-based Action Chunking Policies}

Imitation learning has become a mainstream paradigm for robotic visuomotor control \supercite{b1,b2,b3,b9}. Early behavior cloning (BC) methods suffered from distribution shift due to frame-by-frame action prediction \supercite{bDAgger,b9}, whereas action chunking improves temporal consistency and robustness in ACT-style policies \supercite{b1}. ACT \supercite{b1} uses a CVAE-Transformer \supercite{bCVAE} architecture to model robot control as joint prediction of a fixed-length future action sequence and achieves strong performance in fine-grained bimanual manipulation. The action chunking mechanism provides temporal abstraction: instead of committing to a single-step action, the policy commits to a trajectory segment, which can smooth execution and reduce compounded errors \supercite{b1}.

Subsequent work has extended this line in several directions. Diffusion Policy \supercite{b2} replaces the CVAE formulation with a diffusion process, requiring multi-step iterative sampling at inference time but representing more complex multi-modal action distributions. BeT \supercite{b3} introduces discrete action representations by discretizing the action space, providing an intermediate design between continuous and discrete control. More recently, large-scale VLA models such as RT-2 \supercite{b18} and $\pi$0.5 \supercite{b10} have shown the potential of extending action chunking through web-scale pretraining and broad task generalization. Concurrently, Implicit Behavioral Cloning (IBC) \supercite{b17} approaches action distribution modeling from an energy-based perspective. Nevertheless, these imitation-learning methods fundamentally depend on the quality of offline demonstrations and do not improve through online task reward, which limits their ability to handle distribution shift \supercite{bDAgger,b9}.

\subsection{Chunk-level Reinforcement Learning}

Early work combining action chunking with RL mainly addressed high-dimensional action spaces through discrete action representations. CQN-AS \supercite{b7} discretizes the action chunk space into a prototype library and learns a Q-function over it, but discretization inherently limits the precision of continuous control \supercite{b7}. This precision tradeoff is especially undesirable for contact-rich manipulation tasks that require fine force modulation.

Recent studies have explored continuous action chunk RL through different approaches. Chunking the Critic \supercite{b5} extends SAC \supercite{b15} to the chunk level by introducing a lightweight Transformer for joint value estimation over action sequences, while retaining a step-wise actor. Q-Chunking \supercite{b6} uses a flow-matching framework for continuous chunk generation but relies on large-scale offline datasets and computationally intensive distillation, limiting its applicability in low-data settings. Concurrent work AC3 \supercite{b14} introduces asymmetric actor updates for continuous chunk RL under sparse rewards and reports promising results across several manipulation benchmarks.

However, these studies mainly focus on learning chunking policies from scratch or optimizing offline-to-online transfer \supercite{b5,b6,b14}. The distinct problem of fine-tuning pretrained action chunking policies, where the goal is to selectively improve task performance through RL while preserving the behavior structure learned from demonstrations, remains insufficiently studied. This is the gap addressed by our work. It is also worth noting that the above studies differ fundamentally from ours in task setting, simulation environment, reward design, and the source of the initial policy, and therefore do not constitute direct quantitative baselines.

\subsection{RL Fine-tuning of Pretrained Policies}

RL fine-tuning of pretrained policies has become a promising paradigm for robotic manipulation \supercite{b14,b20,b21,bDPPO}. The central insight is that BC provides a behavior prior, and RL can subsequently optimize this prior through environmental interaction \supercite{b14,b20,b21}. This approach is particularly attractive for industrial applications because (a) a moderate number of demonstrations may be available, (b) task specifications may change or require customization, and (c) large-scale random exploration is often infeasible under safety constraints \supercite{bSafeLearning}.

Existing work follows two main routes. The first adapts or post-trains large-scale VLA models to exploit their broad generalization ability \supercite{b18,b23,b10}. However, large-scale VLA models typically introduce a vision-language backbone and an action generation module, making their inference path more complex than that of pure vision-action policies. For example, the $\pi$0.5 model used in this paper is built on a PaliGemma vision-language backbone and an action expert; at inference time, it first produces high-level semantic subtasks and then generates continuous action chunks with the action expert \supercite{b10}. This structure supports open-world generalization but also makes inference latency and GPU memory usage important considerations in high-frequency closed-loop control. The second route focuses on generative policy frameworks such as DPPO \supercite{bDPPO}, which fine-tunes Diffusion Policy through a differentiable reward-weighted denoising process. Although effective, the iterative sampling nature of diffusion models introduces additional latency and training variance.

In contrast, deterministic ACT-style action chunking policies generate future action chunks with a single forward pass \supercite{b1}. The original ACT system reports an inference time of approximately 0.01s for an 80M-parameter model on a single RTX 2080 Ti \supercite{b1}. Owing to their simple architecture, temporally continuous actions, and lower deployment overhead, such policies are suitable backbones for industrial RL fine-tuning \supercite{b1}. This paper shows that, with an appropriate MDP reformulation and constraint mechanism, these policies can be effectively fine-tuned by RL to improve task performance while preserving deployment-friendly properties.

\section{Method}

\subsection{Problem Definition: Structural Mismatch}

Standard RL algorithms are formulated around step-wise decisions: at each time step $t$, the policy observes $s_t$, outputs $a_t$, and receives $r_t$ \supercite{b4,b15,b16}. In contrast, ACT uses a CVAE-Transformer architecture to generate a continuous $K=100$-step action chunk $\mathbf{A}_{t} = (a_t, a_{t+1}, \ldots, a_{t+K-1})$ at once, and then executes only the first $c_0=10$ steps under temporal ensembling before replanning \supercite{b1}. During RL fine-tuning, we set the execution horizon to $c=8$, forming the basic temporal unit for chunk-level decision making. The sub-actions inside each chunk exhibit strong temporal coupling and trajectory continuity: each action is not an independent primitive but part of a coordinated motion sequence.

This structural difference introduces three interrelated challenges for step-wise RL fine-tuning. First, the probability ratio $\pi_\theta(a_t|s_t) / \pi_{\theta_{\text{old}}}(a_t|s_t)$ used in the PPO clipped surrogate objective operates at the per-action level \supercite{b4}, whereas individual actions inside a chunk are not independently meaningful: their joint structure is the key object. Computing ratios for individual actions assigns credit and penalties to atomic actions that were never designed to be evaluated independently. Second, the advantage estimate $\hat{A}_t$ is computed per time step, but the reward signal for successfully executing an action chunk may only become visible after multiple steps. This temporal mismatch makes step-wise advantage estimation poorly aligned with the actual credit assignment structure. Third, exploration in the generated $K \times d_a$-dimensional action space is intrinsically harder than exploration in the $d_a$-dimensional single-step action space. We refer to these issues collectively as the ``structural mismatch'' between action chunk generation and step-wise RL optimization.

\begin{figure*}[!t]
\centering
\includegraphics[width=\textwidth]{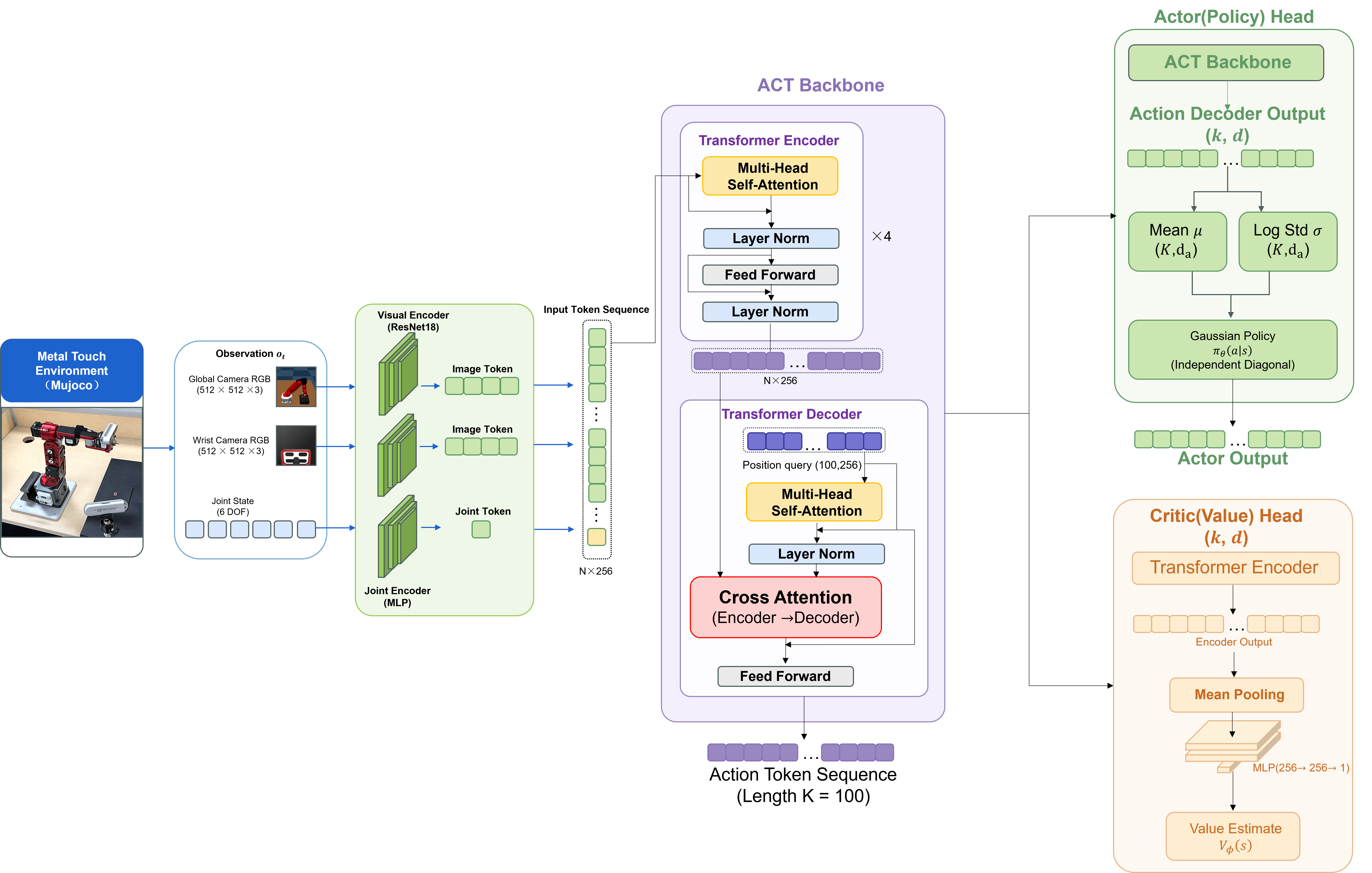}
\caption{Overall network architecture of PAC-ACT. The PPO post-training framework is built from a pretrained ACT model. The Actor branch preserves the original ACT action chunk generation structure while removing the CVAE module, producing temporally consistent continuous action sequences. The Critic branch reuses the ACT visual encoder and Transformer encoder, removes the action decoder, pools the encoder output, and maps it to a scalar state value through a value head.}
\label{fig:architecture}
\end{figure*}

\subsection{Chunk-level MDP Adaptation}

To address structural mismatch, we reformulate the standard step-wise MDP as a chunk-level decision process. The core idea is to group $c$ environment steps into one chunk-level time step, so that policy action generation and RL optimization operate at the same temporal granularity.

Formally, we define a chunk-level MDP in which each chunk time step $\tau$ corresponds to $c$ original environment steps. At chunk time step $\tau$, the policy observes the current state $s_\tau$ (i.e., $s_{\tau c}$ in the original step index) and outputs a $c$-step action chunk $\pi_\theta(\mathbf{a}_\tau | s_\tau)$, obtained by taking the first $c$ actions from the $K$ actions generated by ACT. The environment executes all $c$ actions sequentially and accumulates the reward:
\begin{equation}
R_\tau = \sum_{i=0}^{c-1} r_{\tau c + i},
\end{equation}
and transitions to $s_{\tau+1} = s_{\tau c + c}$. The inter-chunk discount factor is adjusted accordingly as $\Gamma = \gamma^c$, where $\gamma=0.99$ is the original step-wise discount factor.

Under this formulation, the components of PPO naturally adapt to the chunk level. The probability ratio is computed over the entire action chunk:
\begin{equation}
\rho_\tau(\theta) = \frac{\pi_\theta(\mathbf{a}_\tau | s_\tau)}{\pi_{\theta_{\text{old}}}(\mathbf{a}_\tau | s_\tau)},
\end{equation}
where $\pi_\theta(\mathbf{a}_\tau | s_\tau)$ is defined over the flattened $c \times d_a$-dimensional action space. We model the policy with a diagonal Gaussian distribution, so $\log\pi_\theta(\mathbf{a}_\tau | s_\tau)$ decomposes into the sum of log probabilities over action dimensions, and the corresponding probability ratio is the product of dimension-wise ratios. This decomposition provides a consistent stochastic estimate of $\rho_\tau(\theta)$, while the mean $\mu_\theta(s_\tau)$ is still generated by the ACT decoder through self-attention, preserving temporal correlations among steps within the action chunk. Generalized advantage estimation (GAE) is computed at the chunk level:
\begin{equation}
A_\tau^{\text{GAE}} = \sum_{l=0}^{\infty} (\Gamma\lambda)^l \delta_{\tau+l}, \quad \delta_\tau = R_\tau + \Gamma V(s_{\tau+1}) - V(s_\tau),
\end{equation}
where $\Gamma = \gamma^c$ is the inter-chunk discount factor.

In implementation, we use a hybrid scheme of step-wise reward collection and chunk-level policy update. Rewards are collected at every environment step to retain fine-grained feedback, while GAE advantages are aggregated at chunk boundaries and the PPO clipped surrogate loss is computed over complete chunks. This preserves detailed reward information while aligning the optimization granularity with the action generation structure.

\subsection{ACT-transferred Actor-Critic}

\subsubsection{Actor: Preserving Action Chunk Generation}

In this architecture (Figure~\ref{fig:architecture}), the Actor fully reuses the visual encoder (ResNet-18 \supercite{b11}), Transformer \supercite{b12} encoder (4 layers, 256 dimensions), and action decoder of the pretrained ACT model to preserve the temporal action structure learned during pretraining. The Actor input is a multimodal observation consisting of two RGB images (front and wrist views, 512$\times$512) and a 6-dimensional joint-angle encoding. The visual encoder processes each camera view independently into latent features, which are concatenated with low-dimensional proprioceptive signals. The Transformer encoder then processes the fused representation with self-attention to capture spatial and temporal relations. Finally, the action decoder outputs a $K \times d_a$ tensor ($K=100$), representing the action mean at each step over the generated horizon.

The original ACT uses a CVAE to model action diversity through a latent variable $z$, sampled from the prior distribution at inference time \supercite{b1,bCVAE}. During RL fine-tuning, the policy already requires stochastic exploration through its Gaussian policy distribution; retaining the CVAE would introduce an additional and redundant source of randomness, increasing optimization instability. More importantly, the stochastic latent variable $z$ in the CVAE conflicts with the baseline penalty in our hybrid KL constraint (Eq.~\eqref{eq:reward_baseline}): $z$ induces inherent output fluctuations, while the $\beta_2$ term continuously penalizes these fluctuations, canceling useful gradients and reducing optimization efficiency. The ablation study in Sec.~\ref{sec:ablation} quantitatively confirms that retaining the CVAE leads to a measurable degradation in convergence. In addition, recent progress in visual representation learning, such as R3M \supercite{b19}, provides rich pretrained initialization options for robotic visual encoders. Based on this analysis, we remove the CVAE module from the Actor and use the network output directly as the mean of a diagonal Gaussian policy:
\begin{equation}
\mu_\theta(s) = \text{ACT}_{\text{backbone}}(s), \quad a \sim \mathcal{N}(\mu_\theta(s), \Sigma),
\end{equation}
where the covariance matrix $\Sigma = \text{diag}(\sigma_1^2, \ldots, \sigma_{c \cdot d_a}^2)$ is represented by learnable log-standard-deviation parameters, initialized as $\log \sigma_i = -5.5$ for Metal Touch or $-3.0$ for Square Assembly to match the output scales of different action representations and encourage initially narrow exploration around the pretrained behavior. This parameter is selected according to the action representation of the pretrained policy (absolute position vs. delta action), reflecting the adaptability of the framework to different action spaces.

\subsubsection{Critic: Lightweight Encoder-Value Network}

The Critic maps high-dimensional observations to a scalar state value $V(s)$. We construct the Critic by reusing the visual encoder and Transformer encoder of the pretrained ACT model, removing the action decoder, and appending a value head. The Critic and Actor use the same visual encoder and Transformer encoder architecture, except that the Critic has no decoder and maintains an independent copy of the weights. Given an observation $s$, the Critic runs the visual encoder and Transformer encoder, mean-pools the output sequence representation into a global encoded feature vector, and feeds it into a lightweight value network:
\begin{equation}
V(s) = \text{MLP}(\text{pool}(\text{ACT}_{\text{encoder}}(s))),
\end{equation}
where the MLP head contains two hidden layers (256$\rightarrow$256$\rightarrow$1) with ReLU activations. Removing the decoder is motivated by the following consideration: the Critic estimates state value rather than generating actions, and the additional parameters introduced by the decoder do not directly contribute to value estimation. The hidden representation produced by the encoder already contains task-relevant visual and state information; after pooling, it can be used effectively for value regression while reducing the Critic's parameter count and computational cost. The ablation study in Sec.~\ref{sec:ablation} further shows that a Critic retaining the decoder lags behind by approximately 12 percentage points at the same training budget, validating the encoder-value design.

\subsection{Hybrid KL-Regularized Policy Optimization}

Standard PPO \supercite{b4} uses a clipped surrogate objective to constrain policy updates:
\begin{equation}
\mathcal{L}_{\text{PPO}} = \mathbb{E}_t\Big[ \min\big( r_t(\theta) \hat{A}_t,\; \text{clip}(r_t(\theta), 1-\epsilon, 1+\epsilon) \hat{A}_t \big) \Big],
\end{equation}
where the probability ratio is defined as
\begin{equation}
\rho_t(\theta) = \frac{\pi_\theta(a_t|s_t)}{\pi_{\theta_{\text{old}}}(a_t|s_t)}.
\end{equation}

However, in the high-dimensional action chunk space, the clipped objective alone is insufficient to prevent destructive drift away from the pretrained behavior manifold. We therefore introduce two complementary regularization mechanisms.

First, we add a KL penalty \supercite{b16} to the loss function to constrain distributional drift between adjacent PPO updates:
\begin{equation}
\mathcal{L}_{\text{train}} = -\mathcal{L}_{\text{PPO}} + \beta_1 \cdot \widehat{D}_{\text{KL}}(\pi_{\theta_{\text{old}}} \| \pi_\theta) + \ldots,
\end{equation}
where $\beta_1 = 3.0$. In implementation, we use the empirical KL estimate between adjacent PPO policies as the regularization term and combine it with the PPO clipped objective to constrain distributional drift. This local trust-region constraint complements the clipped surrogate objective by providing a softer but more global penalty. It is particularly important in the action chunk space, where per-action clipping boundaries are insufficient to control joint distributional drift across all $c \times d_a$ dimensions.

Second, we introduce behavior-prior regularization at the reward level. Using the pretrained ACT baseline model $\pi_{\text{base}}$ with fixed parameters, we penalize deviations between the current policy output and the baseline output during rollout collection:
\begin{equation}
r'_t = r_t - \beta_2 \cdot \| \mathbf{a}_\theta(s_t) - \mathbf{a}_{\text{base}}(s_t) \|^2,
\label{eq:reward_baseline}
\end{equation}
where $\beta_2 = 2.0$. The modified reward $r'_t$ replaces the original $r_t$ in GAE computation. This mechanism encourages RL updates to explore within the neighborhood of the expert action manifold and prevents the policy from drifting into regions of the action space that may be assigned high value by the learned value function but would not be executed by the pretrained policy and may be unsafe or unstable.

\begin{algorithm}[htbp]
\small
\caption{PAC-ACT (Chunk-level PPO Fine-tuning)}
\label{alg:actppo}
\begin{algorithmic}[1]
\State \textbf{Input:} pretrained ACT policy $\pi_\theta$, initial covariance $\Sigma$, value network $V_\phi$, rollout buffer $\mathcal{B}$
\State \textbf{Hyperparameters:} chunk execution length $c$, PPO epochs $E$, minibatch size $M$, number of parallel environments $N$, $\beta_1$, $\beta_2$
\State Load baseline policy $\pi_{\text{base}} \gets \pi_\theta$\hfill\text{// Frozen and not updated}
\For{iteration $=1,2,\ldots$}
    \State \textbf{Stage 1: Rollout collection}
    \For{each environment $i=1,\ldots,N$}
        \State $s \gets \text{env.reset}()$
        \For{$t = 1$ \textbf{to} rollout\_steps}
            \State $\boldsymbol{\mu}, v \gets \pi_\theta(s), V_\phi(s)$
            \State $\mathbf{a} \sim \mathcal{N}(\boldsymbol{\mu}, \Sigma)$\hfill\text{// Output a $c$-step action chunk}
            \State $s', r_{\text{env}}, \textit{done} \gets \text{env.step}(\mathbf{a})$\hfill\text{// Execute $c$ internal substeps}
            \State $d = \| \boldsymbol{\mu} - \boldsymbol{\mu}_{\text{base}}(s) \|^2$, $r = r_{\text{env}} - \beta_2 d$
            \State $\mathcal{B} \gets (s, \mathbf{a}, r, v)$, $s \gets s'$
        \EndFor
    \EndFor
    \State \textbf{Stage 2: Chunk-level advantage estimation}
    \State Sample from $\mathcal{B}$ and compute GAE:
    \State \quad $\delta_t = r_t + \Gamma V_\phi(s_{t+1}) - V_\phi(s_t)$
    \State \quad $A_t^{\text{GAE}(\Gamma,\lambda)} = \sum_{l=0}^{\infty} (\Gamma\lambda)^l \delta_{t+l}$
    \State \textbf{Stage 3: Chunk-level PPO update}
    \For{epoch $=1$ \textbf{to} $E$}
        \For{minibatch $=1$ \textbf{to} $M$}
            \State $\rho_t = \pi_\theta(\mathbf{a}_t|s_t) / \pi_{\theta_{\text{old}}}(\mathbf{a}_t|s_t)$
            \State $\mathcal{L}_{\text{PPO}} = \min(\rho_t A_t,\; \text{clip}(\rho_t, 1-\epsilon, 1+\epsilon)A_t)$
            \State $\mathcal{L}_{\text{value}} = \text{MSE}(V_\phi(s_t),\; V_\phi^{\text{target}}(s_t))$
            \State $\mathcal{L} = -\mathcal{L}_{\text{PPO}} + \beta_1 \widehat{D}_{\text{KL}}(\pi_{\theta_{\text{old}}}, \pi_\theta) + c_v\mathcal{L}_{\text{value}} - c_e\mathcal{H}(\pi_\theta)$
            \State Update $\theta, \phi$
        \EndFor
    \EndFor
    \If{converged, exit}
    \EndIf
\EndFor
\end{algorithmic}
\end{algorithm}

\section{Experiments}

\subsection{Metal Touch Precision Contact Scenario}

\begin{figure}[!tbp]
\centering
\includegraphics[width=0.95\columnwidth]{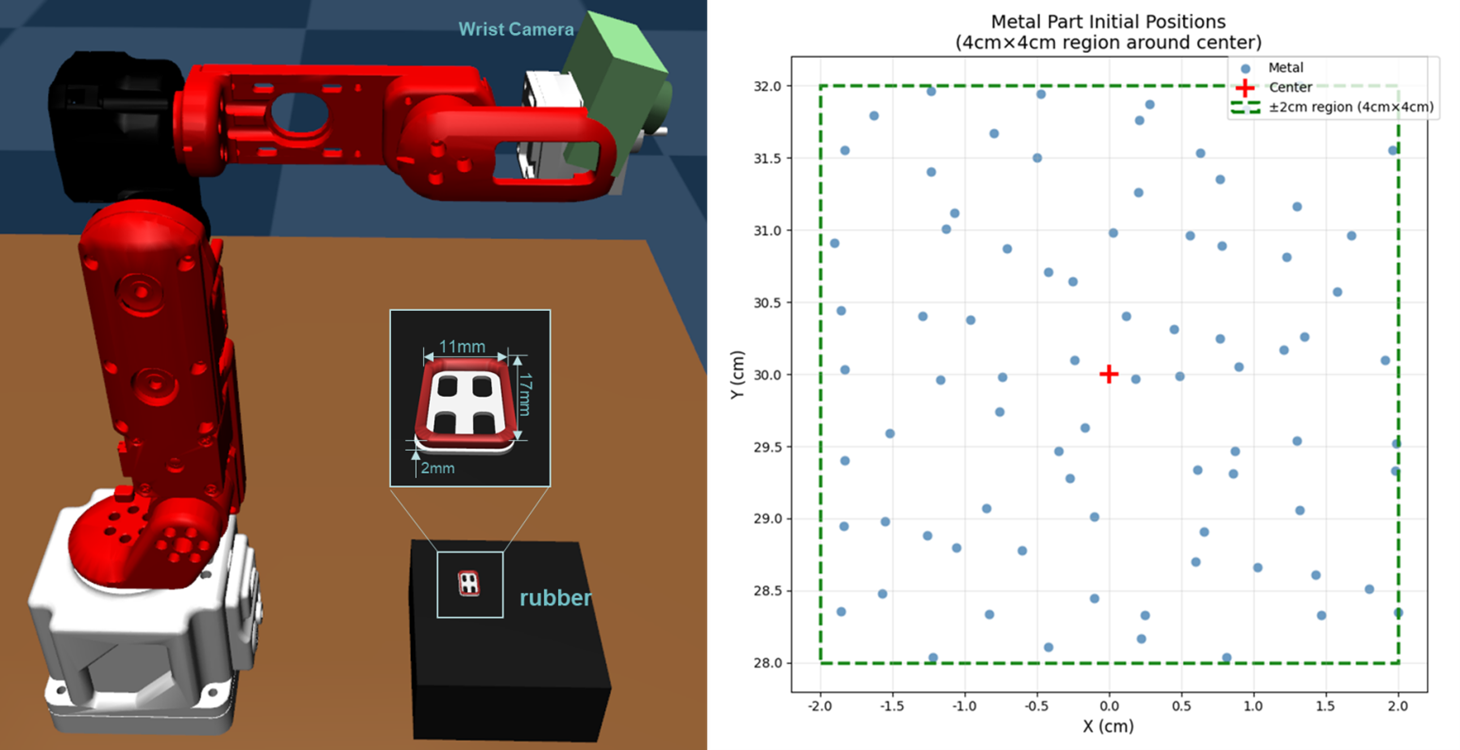}
\caption{Metal Touch precision contact task environment and randomized initialization distribution. Left: continuous contact manipulation in MuJoCo \supercite{b13}, where a 6-DoF robotic arm performs precision contact operations on a rubber-metal component through an end-effector probe. Right: randomized initial positions of the metal component across episodes; the object is randomly shifted within $\pm$2cm around the center region to simulate positioning errors and out-of-distribution perturbations in industrial scenarios.}
\label{fig:environment}
\end{figure}

Metal Touch is a simulated benchmark for precision contact manipulation. As shown in Figure~\ref{fig:environment}, the task scene contains a 6-DoF robotic arm equipped with a stick-like end-effector and a carrier platform fixed on the table (100mm $\times$ 100mm $\times$ 50mm). An annular rubber-metal part is placed on the platform, and the rubber boundary corresponds to four contact regions. The task objective is for the robot to touch all four regions sequentially; the contact force threshold [0.5N, 60.0N] is used for reward constraints and force-safety evaluation.

At the beginning of each episode, the rubber component is randomly translated on the table plane within $\pm$2cm. The policy observation includes two Intel RealSense D415 RGB cameras (front and wrist views, 512$\times$512 resolution) and a 6-dimensional joint-angle encoding. The MuJoCo environment also provides 6-axis force/torque sensor readings for reward computation and force-safety evaluation during RL post-training, but these readings are not used as inputs to the pretrained policy. The control frequency is 10Hz, and each RL training episode is limited to 512 steps, approximately 51 seconds.

We design three contact tasks with progressively increasing complexity, covering different trajectory geometries and motion patterns: (1) Diamond, which sequentially touches four side points; (2) Cross, which follows center $\to$ left $\to$ center $\to$ right $\to$ center $\to$ up $\to$ center $\to$ down and contains observation-action ambiguity; and (3) Contour, which continuously moves along the inner boundary of the rubber ring and has the longest effective contact horizon. These tasks differ in trajectory shape, motion direction, and contact timing, jointly testing policy adaptability across diverse contact scenarios. Task success is measured by whether all four regions are reached, and whether the contact force exceeds 60N is reported as an independent force-safety metric. We additionally test scalability on robomimic \supercite{b9} Square Assembly.

\subsection{Experimental Setup}

\subsubsection{Dataset Construction}

All demonstration data are collected and stored using the LeRobot framework. LeRobotDataset adopts a hierarchical storage architecture: high-frequency low-dimensional signals, including joint angles, force sensor readings, and action commands, are organized in Apache Parquet columnar format, with each row corresponding to one frame; visual observations, including front and wrist RGB images, are temporally compressed into camera-specific MP4 files. The metadata layer records the schema, frame rate, normalization statistics, and episode boundaries. This design minimizes disk storage through video compression while enabling efficient random access to individual frames during training. To study the effect of dataset size, we construct datasets containing 50, 100, 200, 400, and 800 expert trajectories, retaining only successful trajectories. For Square Assembly, robosuite \supercite{bRoboSuite} replay is used to reconstruct the data in LeRobotDataset format.

\subsubsection{PPO Training Configuration and Evaluation Metrics}

The pretrained ACT model uses a ResNet-18 visual encoder, a 4-layer Transformer encoder with 256-dimensional hidden states, and an action generation length of $K=100$ with temporal ensembling execution length $c_0=10$. It is pretrained on 800 trajectories for 200,000 steps. PPO hyperparameters are as follows: chunk execution length $c=8$, $\gamma=0.99$, GAE $\lambda=0.95$, clipping range $\epsilon=0.2$, learning rate $1\times10^{-6}$, Critic learning rate $1\times10^{-5}$, entropy coefficient 0.001, KL coefficients $\beta_1=3.0,\ \beta_2=2.0$, PPO epochs 8, minibatch size 32, and 16 parallel environments. The chunk execution length is adjusted from the pretrained $c_0=10$ to $c=8$ because Metal Touch uses an absolute-position action representation, making the policy more sensitive to the initial value of $\log\sigma$. Shortening the executed chunk reduces training difficulty while maintaining exploration entropy and effectively preserves the success-rate basis of the pretrained $c_0=10$ temporal ensembling policy.

The reward function follows a milestone design. A region-touch reward is triggered when a new region is contacted for the first time. Additional penalty terms include a distance penalty to suppress unnecessary end-effector displacement, a smoothness penalty for abrupt actions, an orientation penalty to maintain consistent end-effector pose, a time penalty to encourage efficient completion, and a force-control penalty when the contact force exceeds the safety range. A success bonus is given when all four regions are successfully reached. The complete environment reward is:
\begin{equation}
\begin{split}
r_t ={}& r_{\text{touch}} - p_{\text{dist}} - p_{\text{smooth}} - p_{\text{orient}} \\
& - p_{\text{time}} - p_{\text{force}} + r_{\text{success}},
\end{split}
\label{eq:env_reward}
\end{equation}
Here $p_{\text{force}}$ is triggered when the total contact force $f_{\text{total}}$ exceeds the safety range. All main experiments use the dense reward in Eq.~\eqref{eq:env_reward}; the sparse-reward ablation in Sec.~\ref{sec:ablation} retains only the success reward. This $r_t$ is the original reward in Eq.~\eqref{eq:reward_baseline}, which is modified by the baseline penalty to obtain the final PPO reward $r'_t = r_t - \beta_2 d$.

Evaluation uses deterministic rollouts without exploration noise and reports statistics over 50 episodes. During training, the policy introduces stochastic exploration through a Gaussian distribution ($\log\sigma=-5.5$); during evaluation, the policy mean is executed to avoid sampling noise in performance estimation. In addition, the object initial position is randomized within $\pm$2cm in each episode to evaluate execution stability under common positioning errors. Unless otherwise stated, RL training and key validations are repeated with three random seeds (22, 32, 42). Key metrics include task success rate, completion steps, and force-safety statistics, including peak contact force and the proportion of force readings above safety thresholds. All RL training and evaluation are conducted on an NVIDIA A100-SXM4-80GB GPU.

\subsection{Multi-policy Success Rate Comparison}

We compare ACT, Diffusion Policy, $\pi$0.5 \supercite{b10}, a 3B-parameter VLA foundation model, and PAC-ACT under a unified environment. All pretrained models are trained on 800 episodes of data (ACT: 200k steps, Diffusion Policy: 200k steps, $\pi$0.5: 20k steps). Evaluation is performed on the Contour task and Square Assembly. Table~\ref{tab:success} reports the success rates for both tasks.

\begin{table}[!t]
\caption{Success rate comparison across policies.}\label{tab:success}
\begin{center}
\begin{tabular}{lcc}
\toprule
Method & Contour & Square Assembly \\
\midrule
ACT & 60.0\%$^\dagger$ & 51.2\% \\
Diffusion Policy & 60.0\%$^\dagger$ & 77.8\% \\
$\pi$0.5 & 79.0\% & 62.6\% \\
\textbf{PAC-ACT (Ours)} & \textbf{100.0\%} & \textbf{98.2\%} \\
\bottomrule
\end{tabular}
\end{center}
\vspace{2pt}
{\raggedright\footnotesize Metal Touch Contour evaluation uses 50 deterministic rollout episodes, and all methods are tested under the same random seeds. $^\dagger$The identical 60.0\% success rates of ACT and Diffusion Policy on Contour are an experimental coincidence; the two methods are evaluated independently using their own checkpoints pretrained on 800 trajectories.}
\end{table}

After RL fine-tuning, PAC-ACT improves the Contour success rate from 60\% to 100\% and Square Assembly from 51.2\% to 98.2\%. $\pi$0.5 achieves a 79.0\% completion rate on Contour and 62.6\% on Square Assembly, but it exhibits high inference latency variance and large GPU memory usage (44.15GB, Table~\ref{tab:latency}), which creates practical limitations for real-time deployment on edge devices.

\subsection{Online Inference Efficiency Comparison}

The results in Table~\ref{tab:latency} mainly reflect online deployment overhead under the unified implementation and measurement protocol used in this paper. ACT deterministically decodes action chunks \supercite{b1}, and PAC-ACT also performs only an Actor forward pass at deployment, so both methods maintain low mean and tail latency. Diffusion Policy requires multi-step denoising or Langevin sampling to generate actions during inference \supercite{b2}, and its measured mean latency, P95, and P99 are correspondingly higher than those of ACT-style policies. In comparison, $\pi$0.5 combines a PaliGemma vision-language backbone, high-level semantic reasoning, and a flow-matching action expert to generate continuous action chunks \supercite{b10}, resulting in a larger model and a more complex inference chain. The more direct deployment difference lies in resource usage: related efficient VLA work identifies limited compute and GPU memory on robotic platforms as an important bottleneck for deploying large-scale VLAs \supercite{bDeeRVLA}. Consistent with this observation, $\pi$0.5 reaches a peak GPU memory usage of 44.15GB in our test, far above ACT, Diffusion Policy, and PAC-ACT. PAC-ACT achieves a mean inference latency of 88.1ms, approximately 10\% lower than ACT (98.1ms), because removing the CVAE module simplifies the computation graph. Its P95 tail latency of 125.6ms is the lowest among all methods, indicating stable inference latency. When only the Actor is deployed, GPU memory usage is 2.30GB, essentially matching ACT's 2.27GB and making the model size and memory footprint closer to edge-deployment requirements. Real-time deployment on edge devices such as NVIDIA Orin remains to be measured in future work.

\begin{table}[!t]
\caption{Online inference efficiency comparison across policies.}\label{tab:latency}
\centering
\scriptsize
\setlength{\tabcolsep}{3pt}
\begin{tabular*}{\columnwidth}{@{\extracolsep{\fill}}lcccc@{}}
\toprule
Method & Mean (ms) & P95 & P99 & Mem. (GB) \\
\midrule
ACT & $98.1\pm21.7$ & 143.2 & 176.7 & 2.27 \\
Diffusion Policy & $114.9\pm42.2$ & 199.1 & 252.7 & 5.15 \\
$\pi$0.5 & $182.6\pm179.8$ & 653.4 & 739.7 & 44.15 \\
\textbf{PAC-ACT (Ours)} & \textbf{$88.1\pm23.4$} & \textbf{125.6} & \textbf{172.5} & \textbf{2.30}$^\dagger$ \\
\bottomrule
\end{tabular*}

\vspace{2pt}
\begin{minipage}{\columnwidth}
\footnotesize
\raggedright
$^\dagger$ During inference, PAC-ACT loads only the Actor; the Critic is used only for training and is not involved in deployed control. If the Actor-Critic is loaded together, peak GPU memory is 2.59GB.\\
GPU memory is the peak inference-process GPU memory reported by NVML, including model weights, CUDA context, rendering buffers, and other allocations. The peak memory of $\pi$0.5 includes the overhead of the multi-process VLA inference pipeline, of which PyTorch tensor memory is approximately 13.8GB.
\end{minipage}
\end{table}

\subsection{Data Scaling and Performance Bottleneck Analysis}

\subsubsection{Data Scaling Effect of Behavior Cloning}

\begin{figure}[!tbp]
\centering
\includegraphics[width=0.95\columnwidth]{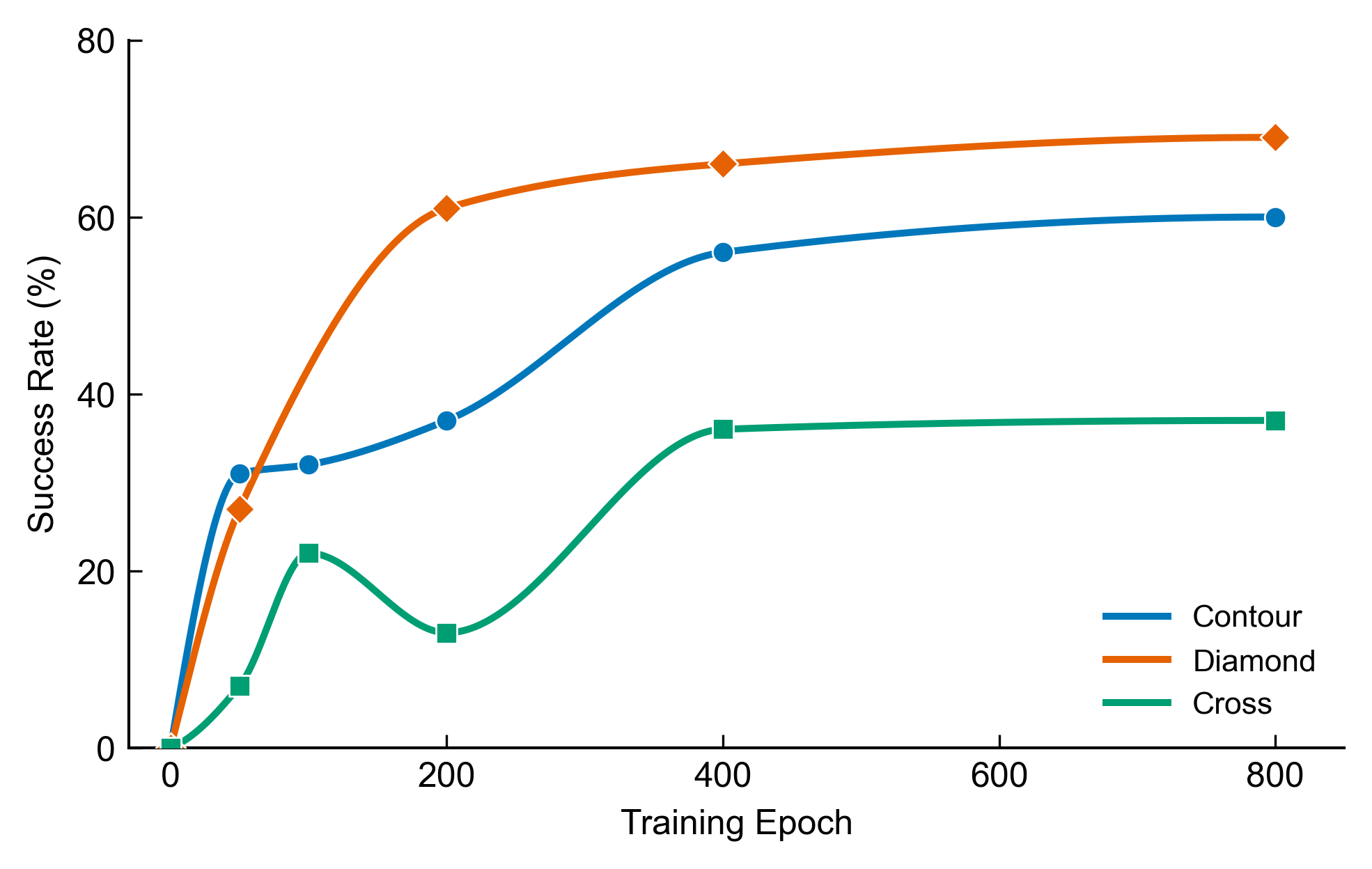}
\caption{Success-rate curves of ACT behavior cloning under different demonstration dataset sizes. As the number of expert demonstrations increases, performance improves on Diamond, Contour, and Cross, but the marginal gain gradually diminishes at larger data scales. The results indicate that pure behavior cloning can learn basic contact trajectories but remains vulnerable to accumulated out-of-distribution errors in long-horizon continuous contact tasks, leading to a performance bottleneck.}
\label{fig:datascaling}
\end{figure}

The behavior cloning success rate of ACT shows clear diminishing returns as the dataset size increases (Figure~\ref{fig:datascaling}): Diamond reaches 69\%, Cross 37\%, and Contour 60\%. Cross contains observation-action ambiguity because the end-effector repeatedly passes through the same center region, and its success rate remains only 37\% even with 800 episodes. When the Contour dataset is expanded to 1600 episodes, the success rate increases to 66\%, only 6 percentage points above the 800-episode baseline. This persistent diminishing return is consistent with prior observations that behavior cloning remains vulnerable to error accumulation under distribution shift \supercite{bDAgger,b9}.

\subsubsection{PAC-ACT Breaks the Performance Bottleneck}

\begin{figure*}[!t]
\centering
\includegraphics[width=\textwidth]{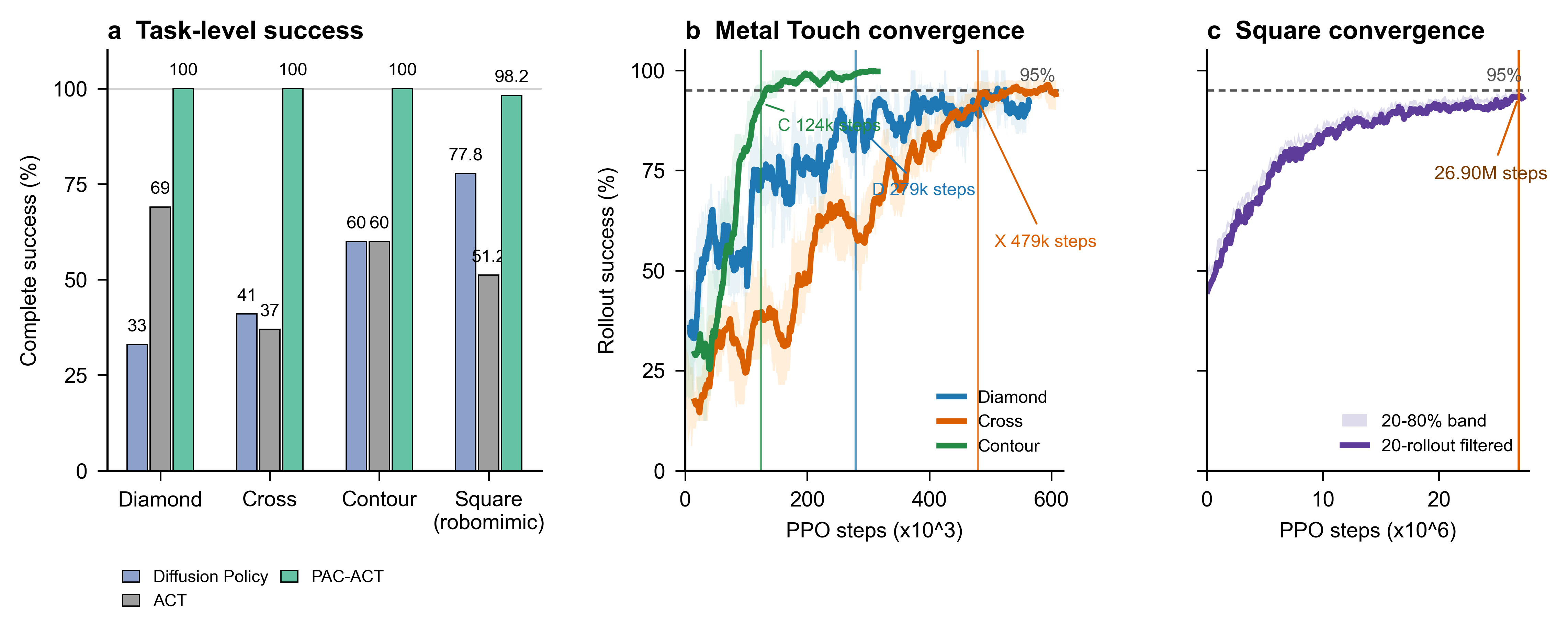}
\caption{Success rates of different policies across manipulation tasks and the convergence process of PAC-ACT. (a) Task-level success rate comparison on Diamond, Cross, Contour, and Square (robomimic), including Diffusion Policy, ACT, and PAC-ACT. (b) PAC-ACT training convergence on Diamond and Contour in Metal Touch; shaded regions indicate local fluctuations in rollout success rate, and the horizontal dashed line marks the 95\% success-rate threshold. (c) PAC-ACT training process on Square (robomimic); the purple curve shows the 20-rollout filtered trend, and the light purple region indicates the corresponding 20--80\% fluctuation range.}
\label{fig:convergence}
\end{figure*}

After applying PAC-ACT to an ACT model pretrained on 800 episodes (Figure~\ref{fig:convergence}), the Contour success rate increases from 60\% to 100\%, and Square Assembly improves from 51.2\% to 98.2\%. This improvement comes from the complementary roles of imitation learning and RL: BC learns basic contact skills and temporal structure from demonstrations, while RL fine-tuning enables the policy to develop error-recovery behavior through environmental interaction \supercite{b14,b20,b21}.

The convergence analysis on Contour shows that PAC-ACT first reaches a 100\% rollout success rate at approximately 100k PPO steps and remains above 95\% from around 150k steps onward. The early-stage success rate is 18.75\% and rises to 72.2\% at approximately 80k steps, indicating a smooth and stable learning process. The final policy maintains a 100\% success rate within approximately 320k total training steps. By contrast, the Diamond task reaches a 95\% rollout success rate at around 250k steps, converging more slowly than Contour and reflecting the higher training difficulty of this precision contact task.

\subsection{Analysis of RL Fine-tuning Effect}

Table~\ref{tab:beforeafter} summarizes the Contour task results before and after fine-tuning.

\begin{table}[!t]
\caption{Comparison before and after PAC-ACT fine-tuning (Contour task, 50-episode evaluation).}\label{tab:beforeafter}
\centering
\begin{tabular*}{\columnwidth}{@{\extracolsep{\fill}}lcc@{}}
\toprule
Metric & ACT (Before RL) & PAC-ACT (After RL) \\
\midrule
Success (4 regions) & 60\% & 100\% \\
Touched regions (/4) & 3.14 & 4.0 \\
Completion steps & 485.4 & 170.6 \\
Episode duration & 48.5s & 17.1s \\
\bottomrule
\end{tabular*}
\vspace{2pt}
{\raggedright\footnotesize During validation, the maximum episode duration is 60s (600 steps), and an episode terminates early after all four regions are reached. This duration limit differs from the 512-step episode limit used during RL training. Validation is repeated with three random seeds (22, 32, 42), and all seeds eventually reach a 100\% rollout success rate.}
\end{table}

After fine-tuning, PAC-ACT achieves an approximately 2.8-fold reduction in completion steps (485.4$\to$170.6) and an approximately 2.8-fold reduction in duration (48.5s$\to$17.1s). Before fine-tuning, ACT typically terminates due to timeout when it fails and reaches only 3.14 regions on average. After fine-tuning, the policy reliably reaches all four regions and completes the task before the time limit. Repeated validation with three random seeds (22, 32, 42) shows that all seeds eventually reach a 100\% rollout success rate, confirming training stability.

We also retain an early reproduction experiment with a changed end-effector probe to examine task-completion adaptation under end-tool geometry variation. In this experiment, the probe radius is increased from 1.5mm to 2.5mm, and the full pipeline from pretraining to RL fine-tuning is reproduced. The policy reaches a final rollout success rate of 99.7\%, with a peak of 100\%, and all 50 deterministic validation episodes successfully reach the four regions, with an average completion length of $68.8\pm20.8$ steps. Because the larger contact radius changes the pressure-equivalent threshold and this experiment uses an earlier reward configuration, we include it only as a supplementary observation of task-completion adaptation under end-tool geometry changes. Force-safety conclusions are based on the main experiments with the unified probe setting and evaluation protocol.

\subsection{Force Safety Analysis}

We conduct a dedicated force-safety evaluation, comparing ACT, Diffusion Policy, $\pi$0.5, and PAC-ACT under identical conditions: the same 50 randomized seeds and a 300-step execution horizon. The force/torque sensor provides 6-axis readings at every control step, from which we compute the total contact force magnitude $f_{\text{total}} = \sqrt{f_x^2 + f_y^2 + f_z^2}$. The results are summarized in Table~\ref{tab:force}.

\begin{table*}[!t]
\caption{Force-safety comparison (Contour task, 50 episodes, 300 steps per episode).}\label{tab:force}
\begin{center}
\small
\begin{tabular}{lcccc}
\toprule
Metric & ACT & Diffusion Policy & $\pi$0.5 & PAC-ACT (Ours) \\
\midrule
Median peak force & 105.40N & 6.76N & 8.35N & 20.74N \\
Mean force ($\mu \pm \sigma$) & $38.67 \pm 94.12$N & $12.31 \pm 60.15$N & $32.99 \pm 221.75$N & $3.92 \pm 4.31$N \\
Maximum peak force & 8452.5N & 3144.6N & 7349.0N & 120.9N \\
Ratio of force readings $>$30N & 15.0\% & 7.6\% & 13.8\% & 0.3\% \\
Ratio of force readings $>$60N & 4.6\% & 2.0\% & 5.1\% & 0.1\% \\
Ratio of episodes with force $>$60N & 20\% & 56\% & 88\% & 6\% \\
\bottomrule
\end{tabular}
\end{center}
\vspace{2pt}
{\raggedright\footnotesize Evaluation uses 50 episodes with 300 steps per episode and randomized object initial positions within $\pm$2cm. The force/torque sensor provides 6-axis readings at each control step, and total contact force is computed as $f_{\text{total}} = \sqrt{f_x^2 + f_y^2 + f_z^2}$. The 60N threshold is the safety-force threshold for Metal Touch.}
\end{table*}

BC-trained ACT, Diffusion Policy, and $\pi$0.5 all exhibit notable force-safety issues. ACT has a median peak contact force of 105.4N, almost twice the 60N safety threshold. Although $\pi$0.5 and Diffusion Policy have lower median forces, 8.35N and 6.76N respectively, they still produce hazardous force events above 60N: Diffusion Policy has a 2.0\% ratio of force readings above 60N, with 56\% of episodes containing at least one above-threshold force reading; for $\pi$0.5, the corresponding ratios are 5.1\% and 88\%. These results may be related to transient action fluctuations produced by generative or sampling-based action generation under contact. ACT reaches an extreme maximum force of 8452.5N, corresponding to a collision event in which the policy pushes the end-effector into the rigid table surface under an out-of-distribution state. Because pure vision-action policies rely only on visual observations and joint encodings during inference, they lack force-sensing feedback. In contrast, contact-manipulation and impedance-control studies generally emphasize the need to regulate robot-environment interaction forces \supercite{bContactSurvey,bImpedance}. Therefore, when contact geometry deviates from the demonstration distribution, pure visual behavior-cloning policies may struggle to detect and correct force anomalies in time.

\begin{figure}[!tbp]
\centering
\includegraphics[width=0.95\columnwidth]{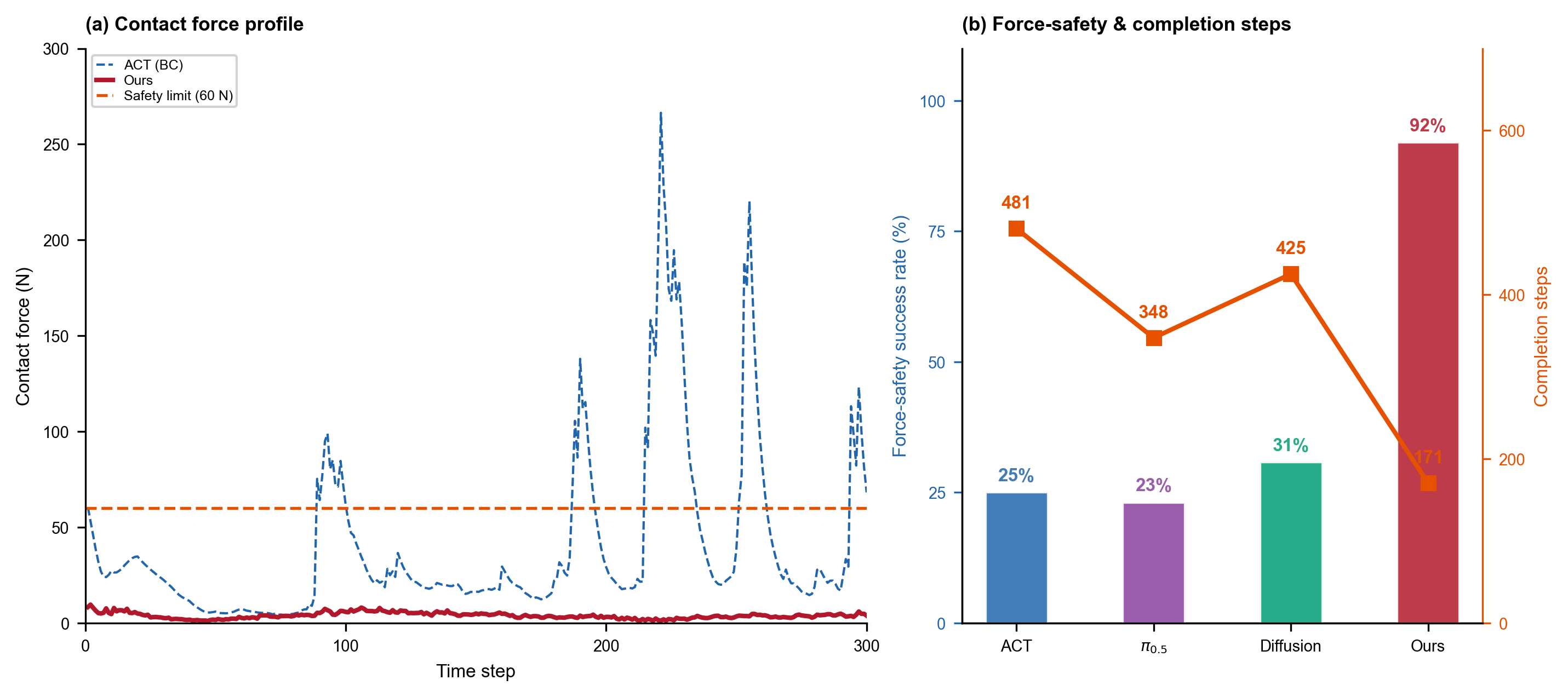}
\caption{Comparison between ACT and PAC-ACT in force control and task-completion efficiency. (a) Contact force distribution comparison; PAC-ACT maintains a more concentrated force distribution and reduces above-threshold hazardous force spikes. (b) Force-safe success rate and completion steps; bars indicate the proportion of successful episodes whose trajectory-level maximum contact force $\max_t f_{\text{total}}$ remains below 60N, and the line indicates average completion steps.}
\label{fig:forcetemporal}
\end{figure}

PAC-ACT effectively improves force safety. Compared with ACT, PAC-ACT reduces the median peak force to 20.74N, below the 60N safety threshold. The mean contact force is also reduced by one order of magnitude (38.67N$\to$3.92N) and is lower than those of Diffusion Policy and $\pi$0.5. Only 0.3\% of force readings exceed 30N, and only 0.1\% exceed 60N, reducing the ratio of force readings above 60N by 46 times relative to ACT. The maximum peak force is reduced by 70 times (8452.5N$\to$120.9N), and the proportion of episodes with any force reading above 60N drops from 20\% for ACT, 56\% for Diffusion Policy, and 88\% for $\pi$0.5 to 6\%. These improvements are associated with the force-safety reward term in the RL objective, which penalizes excessive contact force and encourages gentler and more compliant interaction behavior. Figure~\ref{fig:forcetemporal} compares force control and completion efficiency.

\subsection{Task Completion Efficiency Analysis}

Beyond force-safety metrics, PAC-ACT also improves task success and completion efficiency. Based on Table~\ref{tab:beforeafter} and Figure~\ref{fig:forcetemporal}(b), the Contour task success rate increases from 60\% to 100\%, the average completion length decreases from 485.4 to 170.6 steps, and the average episode duration decreases from 48.5s to 17.1s, a reduction of approximately 2.8 times. Before fine-tuning, ACT often terminates near the time limit and reaches only 3.14 regions on average. After fine-tuning, PAC-ACT reaches all four regions in all 50 deterministic validation episodes and terminates early. These results show that chunk-level PPO post-training not only improves task success but also substantially reduces the interaction steps required to satisfy the success condition.

Qualitative playback of validation videos indicates that the efficiency improvement is not simply uniform acceleration over the entire trajectory, but rather a redistribution of speed across different motion phases. Compared with the policy before fine-tuning, PAC-ACT shows reduced high-frequency end-effector jitter and smoother overall motion. During horizontal or non-contact movements between adjacent contact regions, the end-effector approaches the next target region with larger step sizes, while it slows down and makes smaller adjustments near corners, contact points, or locations requiring precise positioning. This observation is consistent with the quantitative reduction in completion steps: the policy mainly compresses non-critical movement phases while retaining more cautious localization near contact regions. This analysis is based on qualitative video playback and is intended to explain the behavioral manifestation of improved task-completion efficiency, rather than serving as an independent validation of a force-control mechanism.

\subsection{Ablation Study}
\label{sec:ablation}

To validate the Actor-Critic architectural choices, we evaluate two ablation variants:
(1) flat-decoder critic, where the Critic retains the full Transformer decoder and flattens the decoded $K\times d_a$ action chunk sequence before feeding it into the value head;
(2) CVAE-retained actor, where the Actor keeps the CVAE latent module while all other components remain unchanged.
All ablations use exactly the same training hyperparameters as our method, including 16 parallel environments, learning rate $1\times10^{-6}$, 8 PPO update epochs, and $\beta_2=2.0$. Only the corresponding module is changed, ensuring that differences arise from architectural design.

\begin{figure}[!tbp]
\centering
\includegraphics[width=0.85\columnwidth]{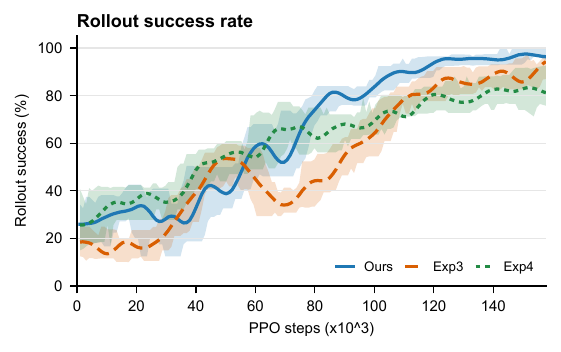}
\caption{Training-process comparison for architecture ablations. Curves show rollout success rate over training steps for the three variants, smoothed with a 10-window moving average.}
\label{fig:ablation}
\end{figure}

Figure~\ref{fig:ablation} and Table~\ref{tab:ablation} show the training process and quantitative comparison of the three variants.

\begin{table}[!t]
\small
\caption{Quantitative comparison of architecture ablations (up to 154K steps).}\label{tab:ablation}
\centering
\begin{tabular*}{\columnwidth}{l@{\extracolsep{\fill}}c@{\extracolsep{\fill}}c}
\toprule
Variant & Final-window mean success$^\dagger$ & Std. dev. \\
\midrule
Ours & 96.4\% & $\pm$3.0\% \\
Flat-decoder critic & 88.0\% & $\pm$7.4\% \\
CVAE-retained actor & 83.1\% & $\pm$7.8\% \\
\bottomrule
\end{tabular*}
\vspace{2pt}
{\raggedright\footnotesize $^\dagger$Mean rollout success rate over the final 20 evaluations.}
\end{table}

The flat-decoder critic retains the full Transformer decoder for value estimation. Experiments show that this variant can also reach a peak rollout success rate of 100\%, indicating that it is not an infeasible architecture. However, its mean over the final 20 evaluations is only 88.0\% with a standard deviation of $\pm$7.4\%, and the success rate fluctuates between 70\% and 100\%, showing insufficient convergence stability. By contrast, our encoder-value Critic achieves a final-window mean of 96.4\% with a standard deviation of only $\pm$3.0\%, maintaining high success while converging smoothly. In addition, the flat-decoder critic retains the full Transformer decoder, introducing approximately 17\% additional parameters and computational overhead (129M$\to$151M parameters), which increases overfitting risk. The hidden representation produced by the encoder already contains sufficient information for state-value estimation, allowing stable value regression without the decoder.

The CVAE-retained actor reaches a final-window mean of only 83.1\% with a standard deviation of $\pm$7.8\%, lower than the 96.4\% of our method. This is because the CVAE latent variable $z\sim\mathcal{N}(0,I)$ makes the same observation produce different action outputs during forward passes, while the baseline MSE penalty $\beta_2\|\mathbf{a}_\theta-\mathbf{a}_{\text{base}}\|^2$ in Eq.~\eqref{eq:reward_baseline} continuously penalizes these inherent stochastic fluctuations. This conflicts with the gradient direction of the PPO objective and lowers the effective signal-to-noise ratio of policy updates. The result indicates that retaining the CVAE causes a measurable final performance loss under the hybrid KL constraint framework.

In summary, the ablation study shows that (1) the encoder-value Critic outperforms the flat-decoder critic in both final-window mean success rate (96.4\% vs. 88.0\%) and training stability ($\pm$3.0\% vs. $\pm$7.4\%) while incurring lower computational overhead; and (2) removing the CVAE helps avoid gradient conflict between an additional source of stochasticity and the KL constraint, as the CVAE-retained actor achieves only an 83.1\% final-window mean, confirming the necessity of CVAE removal under the hybrid KL framework.

To further examine robustness to reward sparsification, we conduct an additional sparse-reward ablation. This ablation is not the main experimental setting: the main experiments use the dense reward in Eq.~\eqref{eq:env_reward}, whereas the sparse ablation retains only the success reward (success\_bonus=12.0) and sets the region-touch reward, distance guidance, smoothness, orientation, time, and force-control terms to zero. Except for the reward, all other settings are the same as the Contour main experiment: the policy is initialized from the same ACT model pretrained on 800 episodes, the initial pose is randomized within $\pm$2cm, the training limit is set to 10000 episodes, and the hybrid KL behavior-prior constraint is retained. No additional demonstration trajectory replay or demonstration-action supervision is used during training.

\begin{figure}[!tbp]
\centering
\includegraphics[width=\columnwidth]{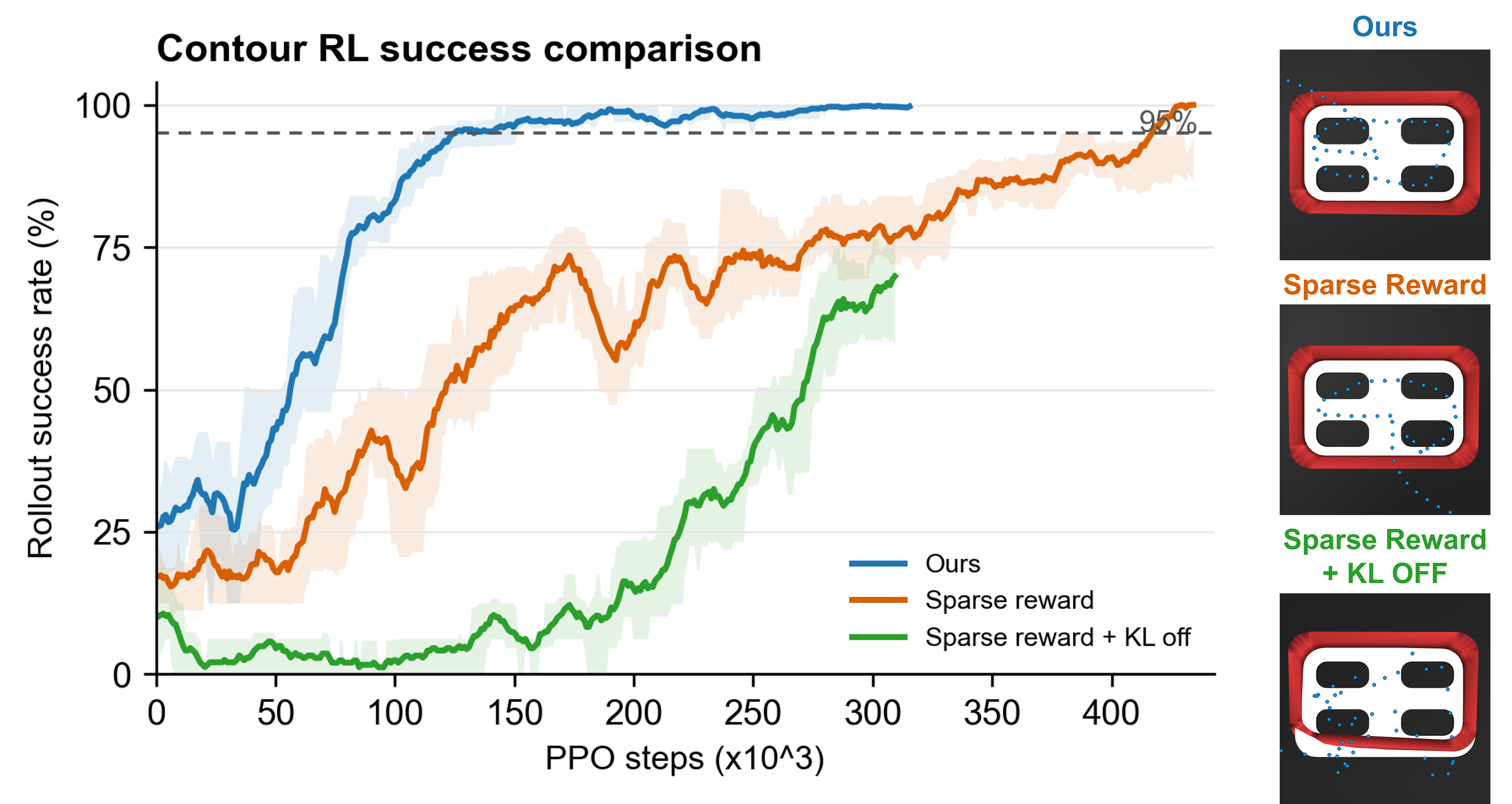}
\caption{Training success rate and end-effector trajectory structure diagnosis in the sparse-reward ablation. Left: rollout success-rate curves under different reward and regularization configurations on the Contour task; shaded regions indicate local fluctuations, and the dashed line marks the 95\% success-rate threshold. Right: top-view trajectory from one validation rollout for the corresponding policy, where blue points visualize the end-effector trajectory sampled uniformly by arc length. When only the success reward is retained but the KL constraint is kept, the policy maintains structured motion close to the contour action sequence. When the KL constraint is removed under the same sparse-reward condition, later rollout success can partially recover, but the sampled end-effector points show a clearly dispersed boundary-touching exploration pattern, indicating that success-rate metrics may mask degradation of temporal structure.}
\label{fig:sparseablation}
\end{figure}

As shown in Figure~\ref{fig:sparseablation}, under this setting PAC-ACT with $\beta_1=3.0$ still converges to a 100\% rollout success rate and maintains a structured trajectory close to the contour action sequence, indicating that the pretrained action chunking policy provides a strong behavior prior for chunk-level PPO. As a comparison, we set the KL coefficient to zero ($\beta_1=0$) under the same sparse-reward setting. Qualitative video playback and end-effector trajectory sampling show that this policy deviates from the pretrained ACT contour temporal action sequence early in training and instead forms shortcut behavior that directly touches multiple target boundaries to obtain the success reward. Because this behavior can still trigger the sparse success condition, the later local rollout window reaches a maximum success rate of 88.5\%; however, under the same training limit, the cumulative total success rate is only 22.0\%, and the final rollout window also drops to 72.7\%. Therefore, the success-rate rebound without KL does not indicate recovery of structured contact skills, but instead shows that relying only on sparse success metrics can obscure degradation of temporal structure. This comparison indicates that KL regularization is a key mechanism for maintaining the pretrained behavior manifold and preventing structured behavior collapse under sparse rewards.

\section{Discussion}

The experiments reveal a paradigm in which imitation learning provides a behavior prior and reinforcement learning performs behavior customization, consistent with the basic idea of recent RL fine-tuning studies on pretrained policies \supercite{b14,b20,b21,bDPPO}. The behavior cloning stage learns basic contact skills and temporal action structure, while the RL stage selectively refines these skills through environmental reward feedback, improving task success, force safety, and time efficiency while preserving the basic behavior structure. The ablations in Sec.~\ref{sec:ablation} further validate the design at the architectural level: the encoder-value Critic outperforms the flat-decoder critic in final-window mean success rate (96.4\% vs. 88.0\%) and stability ($\pm$3.0\% vs. $\pm$7.4\%) with lower computational overhead; removing the CVAE avoids conflict between an additional stochasticity source and the KL constraint, while the CVAE-retained actor achieves only an 83.1\% final-window mean, further supporting the design choice.

Pretrained action chunking policies can reduce sensitivity to reward engineering during RL, but this does not mean that the main method does not require dense rewards. The main experiments use dense rewards to obtain the best task completion efficiency and force safety, while the sparse-reward experiment is only an additional ablation. The results show that, when the KL constraint and frozen baseline policy constraint are retained, keeping only the success bonus still allows convergence under randomized initial poses; when KL is removed, local success may partially recover, but the end-effector trajectory structure clearly degrades. This indicates that the pretrained action chunking policy provides a strong low-level behavior prior, and that the behavior-prior constraint is critical for preventing collapse of structured contact skills under sparse rewards.

The force-safety results deserve particular attention. The substantial reduction in extreme force events, with the peak force decreasing from 8452.5N to 120.9N, shows that a pretrained ACT policy can produce unsafe forces in a non-negligible fraction of episodes even when it achieves a reasonable task success rate. Such events may be missed in pure BC evaluation because (a) they occur in out-of-distribution states that BC evaluation may not systematically probe, and (b) they do not necessarily cause immediate task failure: a policy may generate unsafe force while still completing part or all of the task. RL fine-tuning with an appropriate safety reward provides a principled mechanism for detecting and mitigating such hidden failures, which is relevant to industrial scenarios with strict contact-safety requirements.

This study has several limitations. The experiments are mainly conducted in simulation, and sim-to-real transfer of the learned policy remains to be validated. The current experiments primarily examine object initial-position perturbations and multi-trajectory generalization, but do not cover visual or physical perturbations such as viewpoint changes, lighting variation, or dynamics-parameter changes. A more systematic robustness evaluation is left for future work. Although removing the CVAE module simplifies RL optimization, it may limit the ability of the policy to represent highly multi-modal action distributions. Future work can explore regularized CVAE integration that preserves multimodality while maintaining optimization stability. Finally, the current framework still requires task-specific reward design; developing more automated reward specification methods would further reduce the engineering burden of RL-based policy customization.

\section{Conclusion}

This paper presents PAC-ACT, a chunk-level PPO post-training framework for pretrained ACT policies. Through MDP reformulation, ACT-transferred Actor-Critic design, and hybrid KL constraints, PAC-ACT structurally aligns action chunking policies with RL optimization. On Metal Touch and Square Assembly, the method improves task success (Contour 60\%$\to$100\%, Square 51.2\%$\to$98.2\%) and execution efficiency, reducing completion steps by 2.8 times. Dedicated force-safety analysis shows that hazardous force events are reduced by 46 times and peak contact force by 70 times. All improvements are achieved while maintaining low latency and low GPU memory usage (88.1$\pm$23.4ms inference, 2.30GB GPU memory), indicating the potential of the method for further adaptation and deployment in industrial precision contact scenarios.

Future work will extend the evaluation to real robot platforms to validate sim-to-real transfer, explore more complex multi-stage manipulation tasks, and investigate more general reward design mechanisms that can further reduce the engineering cost of RL-based policy customization.

\section*{Acknowledgment}
This work was supported by the National Natural Science Foundation of China (Grant No. 52675332).

This work was supported by the Regional Joint Fund for Basic and Applied Basic Research Fund of Guangdong Province [grant number
2023B1515120017].

This work was supported bythe Scientific Research Starting Foundation [grant number Y01336136].

This work was supported by High-Level Discipline Construction Fund, Grant/Award Number: G03034K016.

\IEEEtriggeratref{29}

\end{document}